\documentclass[sigconf, nonacm]{acmart}

\newcommand\vldbdoi{XX.XX/XXX.XX}
\newcommand\vldbpages{XXX-XXX}
\newcommand\vldbvolume{19}
\newcommand\vldbissue{1}
\newcommand\vldbyear{2026}
\newcommand\vldbauthors{\authors}
\newcommand\vldbtitle{\shorttitle}
\newcommand\vldbavailabilityurl{https://github.com/FastLM/SPI\_VecDB}
\newcommand\vldbpagestyle{plain} 

\usepackage[T1]{fontenc}
\usepackage{graphicx}
\usepackage{caption}
\usepackage{array}
\usepackage{algorithm}
\usepackage{algorithmic}
\usepackage{tabularx}
\usepackage{amsmath}
\usepackage{amsthm}
\usepackage{booktabs}
\usepackage{subcaption}
\usepackage{multirow}
\usepackage{arydshln}

\usepackage{amssymb}
\usepackage{xcolor}
\usepackage{placeins}
\usepackage{latexsym}
\usepackage{microtype}
\usepackage{textcomp}
\usepackage{mathtools}
\usepackage{paralist}
\usepackage{enumitem}
\usepackage{pifont}

\newtheorem{theorem}{Theorem}

\newcommand{\cmark}{\ding{51}}
\newcommand{\xmark}{\ding{55}}

\begin{document}

\title{SPI: Query-Depth-Adaptive Indexing for Streaming RAG in Vector Databases}

\author{Dong Liu}
\affiliation{%
  \institution{Yale University}
  \city{New Haven}
  \state{CT}
  \country{USA}}
\email{dong.liu@aya.yale.edu}

\author{Yanxuan Yu}
\affiliation{%
  \institution{Columbia University}
  \city{New York}
  \state{NY}
  \country{USA}}
\email{yy3523@columbia.edu}

\begin{abstract}
Vector databases (VecDBs) are increasingly deployed in retrieval-augmented generation (RAG) pipelines where query processing and document ingestion occur concurrently. The index layer needs to provide low-latency search while incorporating new vectors without frequent global rebuilding. Existing VecDB pipelines typically operate within a uniform representation regime, despite substantial variation in the semantic granularity required across queries. This motivates an index design that supports incremental updates while adapting retrieval depth to query distribution and complexity.


We propose \textbf{Semantic Pyramid Indexing (SPI)}, a VecDB-layer indexing framework that organizes embeddings into $L$ semantically aligned resolution levels and selects retrieval depth per query using a lightweight uncertainty-aware controller. SPI supports progressive coarse-to-fine ANN search, level-wise streaming insertion without global rebuilding, and distributed execution through LSH-based partitioning with asynchronous coordination. Unlike hierarchical ANN structures with fixed traversal rules (e.g., SPANN), SPI adapts resolution at query time while remaining compatible with FAISS and Qdrant backends.

On MS MARCO and Natural Questions, SPI achieves competitive Recall@10 with lower latency under the same dense encoder family, yielding a \textbf{1.4--2.3$\times$} average retrieval latency reduction under fixed Recall@10 targets relative to comparable approximate-ANN baselines. A prototype scaling study up to 8 nodes shows $6.2\times$ throughput scaling (${\approx}73\%$ efficiency); the 16-node configuration is included for completeness but shows diminishing efficiency. We provide a top-$K$ stability guarantee: queries with sufficient retrieval margin return an identical top-$K$ set at a shallower level. Code and configurations are available at \url{https://github.com/FastLM/SPI_VecDB}.
\end{abstract}

\maketitle

\pagestyle{\vldbpagestyle}
\begingroup\small\noindent\raggedright\textbf{PVLDB Reference Format:}\\
\vldbauthors. \vldbtitle. PVLDB, \vldbvolume(\vldbissue): \vldbpages, \vldbyear.\\
\href{https://doi.org/\vldbdoi}{doi:\vldbdoi}
\endgroup
\begingroup
\renewcommand\thefootnote{}\footnote{\noindent
This work is licensed under the Creative Commons BY-NC-ND 4.0 International License. Visit \url{https://creativecommons.org/licenses/by-nc-nd/4.0/} to view a copy of this license. For any use beyond those covered by this license, obtain permission by emailing \href{mailto:info@vldb.org}{info@vldb.org}. Copyright is held by the owner/author(s). Publication rights licensed to the VLDB Endowment. \\
\raggedright Proceedings of the VLDB Endowment, Vol. \vldbvolume, No. \vldbissue\ %
ISSN 2150-8097. \\
\href{https://doi.org/\vldbdoi}{doi:\vldbdoi} \\
}\addtocounter{footnote}{-1}\endgroup

\ifdefempty{\vldbavailabilityurl}{}{
\vspace{.3cm}
\begingroup\small\noindent\raggedright\textbf{PVLDB Artifact Availability:}\\
The source code, data, and/or other artifacts have been made available at \url{\vldbavailabilityurl}.
\endgroup
}

\section{Introduction}

Retrieval-Augmented Generation (RAG) pipelines are increasingly deployed in \emph{streaming} settings where queries arrive continuously alongside live document ingestion~\cite{milvus2021milvus}. Production vector databases (VecDBs) such as Milvus, Qdrant, and Weaviate must serve these workloads at low latency while keeping indices fresh---a regime where static, single-resolution indexing structures become a bottleneck: they either over-spend on complex queries or under-serve simple ones.

\begin{figure}
    \centering
    \includegraphics[width=1\linewidth]{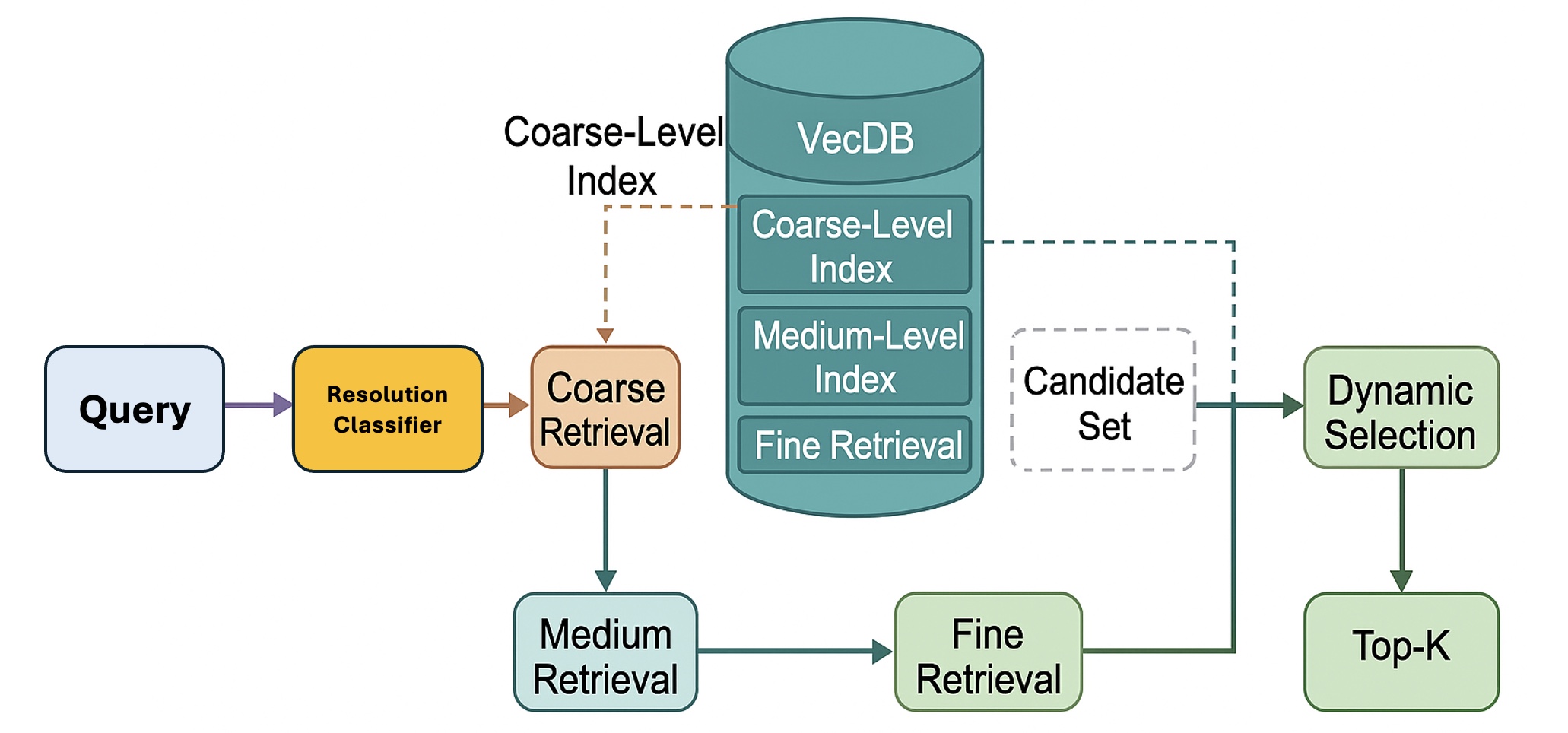}
    \caption{Semantic Pyramid Indexing (SPI) in Vector Databases. The pyramid organizes document embeddings at multiple \emph{representational depth} levels (all levels index the same document corpus; levels differ in semantic discriminability). Incoming streaming queries are routed to the appropriate depth by a lightweight controller, enabling early-exit search for simple queries and deep-representation search for complex ones.}
    \label{fig:spi}
    \vspace{-1em}
\end{figure}

\paragraph{The problem.}
Existing RAG retrieval pipelines commit every query to a single semantic resolution---flat, HNSW, or product-quantized indices embed all documents uniformly and require exhaustive ANN search over a fixed embedding space. This design ignores a fundamental asymmetry in query complexity: a broad query like ``Who discovered gravity?'' is reliably answered by a shallow, high-entropy embedding that excels at broad topical separation, while ``Equation in Newton's third-law paper'' demands a deep, discriminative embedding that can distinguish closely related passages. Here, ``resolution'' refers to the \emph{semantic discriminability} of the embedding space---the ability of the embedding to separate relevant from non-relevant documents---not to the granularity of indexed information units. All pipeline variants index at the document level; the difference is in how much representational depth is devoted to encoding subtle semantic distinctions. Forcing both simple and complex queries through the same single-resolution index wastes compute on the former and may under-resolve the latter. Under streaming workloads, the problem compounds: inserting new documents into a flat or clustered index either blocks live queries (rebuild cost) or degrades index quality (lazy append).

\paragraph{Our approach.}
We propose \textbf{Semantic Pyramid Indexing (SPI)}, a representation-depth-adaptive indexing layer for streaming RAG in vector databases. As shown in Figure~\ref{fig:spi}, SPI constructs an $L$-level semantic pyramid over the \emph{same} document corpus, where level~1 uses shallow broad-scope embeddings and level~$L$ uses deep fine-grained embeddings trained to maximize per-query discriminability via lightweight residual refinement encoders. At query time, a compact uncertainty-aware controller predicts the minimum resolution level sufficient for reliable retrieval, and parallel distributed search is launched across nodes partitioned by the pyramid structure. Documents arriving in the stream are inserted level-by-level without invalidating existing pyramid levels, maintaining semantic consistency under continuous ingestion.

This design makes three concrete contributions over prior work. First, unlike hierarchical indices that use fixed traversal rules, SPI targets a complementary design point: reducing average retrieval cost by routing queries to the minimum representation depth needed for stable top-$k$ retrieval. Second, by co-designing the index layout with distributed parallel execution, SPI achieves throughput scaling that grows proportionally with cluster size up to 8 nodes (79\% efficiency) under bursty streaming arrivals. Third, SPI provides a top-$K$ stability guarantee (Theorem~1): queries whose top-$K$ neighborhood has sufficient retrieval margin retrieve exactly the same top-$K$ set at a shallower level, so early termination is exact for those queries; the margin condition also explains when it is not.

\paragraph{Contributions.}
\begin{itemize}
\item \textbf{SPI framework}: a query-adaptive, multi-resolution indexing framework supporting streaming ingestion in parallel VecDB deployments.
\item \textbf{Theoretical guarantees}: recall preservation and per-query latency bounds under adaptive resolution control with streaming updates.
\item \textbf{Empirical evaluation}: experiments on MS MARCO and Natural Questions show competitive Recall@10 with lower retrieval latency, while streaming and scaling studies evaluate insertion overhead and parallel throughput.
\end{itemize}

\section{Related Work}
SPI sits at the intersection of ANN indexing, streaming vector databases, and retrieval-augmented generation.

\subsection{ANN Indexing in Vector Databases}
Classical ANN foundations include LSH~\cite{andoni2008near}, product quantization~\cite{jegou2011pq,ge2013optimized}, and graph indices such as HNSW~\cite{malkov2018efficient}. Billion-scale systems combine these with pruning and hybrid memory/disk layouts: ScaNN~\cite{guo2020scann}, DiskANN~\cite{jiang2020diskann}, SPANN~\cite{liu2021spann}. GPU-accelerated indices including CAGRA~\cite{ootomo2023cagra} and ParlayANN~\cite{wang2021parlayann} further advance throughput via massive parallelism. Production VecDBs---Milvus~\cite{milvus2021milvus}, Qdrant~\cite{qdrant2021}, Pinecone~\cite{pinecone2021}---integrate these backends with distributed storage and query execution.

\textbf{SPI relationship.} Each SPI pyramid level is a standard FAISS IVF-PQ or Qdrant HNSW index~\cite{douze2024faiss}. SPI adds a semantic resolution layer \emph{above} these backends; it is orthogonal to the specific ANN algorithm used at each level. CAGRA or DiskANN could replace FAISS at any level without changing SPI's control logic.

\subsection{Streaming and Dynamic ANN}
Several recent systems address streaming vector index updates. FreshDiskANN~\cite{subramanya2019freshdiskann} introduced soft deletions and periodic batch consolidation for graph-based ANN. IP-DiskANN~\cite{xu2025ipdiskann} eliminates batch consolidation via in-place insertion/deletion on DiskANN, achieving stable recall without rebuilds. Quake~\cite{mohoney2025quake} takes a complementary approach: adaptive partition scanning that adjusts index structure and query parameters to workload shifts in a NUMA-aware manner, achieving 1.5--38$\times$ lower query latency over HNSW/DiskANN on dynamic workloads. VStream~\cite{gong2025vstream} is a distributed streaming vector search system that introduces hot--cold separation and dynamic partitioning for streaming workloads, achieving 251--373$\times$ query efficiency gains over batch-mode systems.

\textbf{SPI relationship.} These systems address \emph{geometric index freshness}: how to update graph connectivity or partition balance without recall degradation. SPI addresses a \emph{complementary axis}: even given a perfectly fresh index, uniform-resolution ANN still wastes compute on simple queries. SPI's streaming insertion is append-only (no graph repair), trading eventual consistency at finer levels for minimal per-insert latency. The two approaches are directly composable: SPI's level-1 index could be backed by IP-DiskANN or Quake.

\subsection{Retrieval-Augmented Generation}
RAG couples dense or sparse retrievers with generative readers~\cite{lewis2020retrieval,karpukhin2020dense,izacard2021leveraging}. Strong retrievers include ANCE~\cite{xiong2021approximate}, Contriever~\cite{izacard2022contriever}, SPLADE-v2~\cite{formal2022spladev2}, ColBERTv2~\cite{santhanam2022colbertv2}, and HyDE~\cite{gao2022hyde}. End-to-end RAG systems such as Atlas~\cite{izacard2022atlas} jointly train retrievers and readers. Semantic caching~\cite{bang2023gptcache} and FrugalGPT~\cite{chen2023frugal} reduce inference cost but treat retrieval granularity as a pipeline configuration. \textbf{SPI moves query-adaptive resolution control into the VecDB index layer itself.}

\subsection{Comparison with Existing Approaches}
Table~\ref{tab:method-comparison} contrasts SPI with representative systems across key dimensions.

\begin{table}[ht]
  \centering
  \small
  \resizebox{\linewidth}{!}{
    \begin{tabular}{lcccccc}
      \toprule
      \textbf{Method} & \textbf{Multi-Res.} & \textbf{Query-Adap.} & \textbf{Streaming} & \textbf{Semantic} & \textbf{VecDB} & \textbf{GPU} \\
      \midrule
      ColBERTv2     & \xmark & \xmark & \xmark & \cmark & \xmark & \xmark \\
      SPLADE-v2     & \xmark & \xmark & \xmark & \cmark & \xmark & \xmark \\
      SPANN         & \cmark & \xmark & \cmark & \xmark & \xmark & \xmark \\
      IP-DiskANN    & \xmark & \xmark & \cmark & \xmark & \cmark & \xmark \\
      Quake         & \xmark & \cmark & \cmark & \xmark & \cmark & \xmark \\
      CAGRA         & \xmark & \xmark & \xmark & \xmark & \cmark & \cmark \\
      \textbf{SPI (Ours)} & \cmark & \cmark & \cmark & \cmark & \cmark & \cmark \\
      \bottomrule
    \end{tabular}
  }
  \caption{Comparison across key system dimensions. ``Streaming'' = append-capable without full rebuild; ``Query-Adap.'' = per-query traversal depth control; ``GPU'' = GPU-accelerated search.}
  \label{tab:method-comparison}
  \vspace{-1em}
\end{table}

\section{Methodology}

This section presents \textbf{Semantic Pyramid Indexing (SPI)}. To clarify SPI's positioning at the boundary between retriever learning and database systems, we begin with an explicit separation of its two phases.

\paragraph{Two-phase architecture.}
SPI operates in two distinct phases with different system roles:

\begin{itemize}[leftmargin=*,topsep=2pt,itemsep=1pt]
\item \textbf{Phase 1 --- Offline encoder training (retriever learning).} Given a training corpus, learn $L{-}1$ refinement encoders $\{f_\ell, g_\ell\}_{\ell=2}^L$ via joint contrastive training (Section~\ref{sec:training}). This is a \emph{one-time, corpus-specific} cost analogous to fine-tuning DPR or ColBERT. The output is a set of encoder checkpoints.

\item \textbf{Phase 2 --- Online index system (VecDB layer).} Given the trained encoders, build and maintain the $L$-level index $\{\mathcal{I}^{(\ell)}\}_{\ell=1}^L$, serve live queries via progressive retrieval and the controller, handle streaming insertions without rebuilds, and distribute execution across nodes. \emph{This phase is the core database contribution of SPI.}
\end{itemize}

The two phases are \emph{decoupled}: Phase 2 can accept any encoder that produces compatible embeddings---a practitioner who already has a multi-resolution encoder (e.g., from MRL pre-training) can adopt the SPI index without retraining. Conversely, the encoder from Phase 1 could be plugged into a flat FAISS index, but would forfeit the system-level gains of Phase 2. SPI's claimed contributions span both phases, and we evaluate them separately (single-node quality in Q1 reflects Phase 1+2; distributed scaling in Q3 reflects Phase 2 alone).

\begin{figure}[t]
\centering
\includegraphics[width=0.8\linewidth]{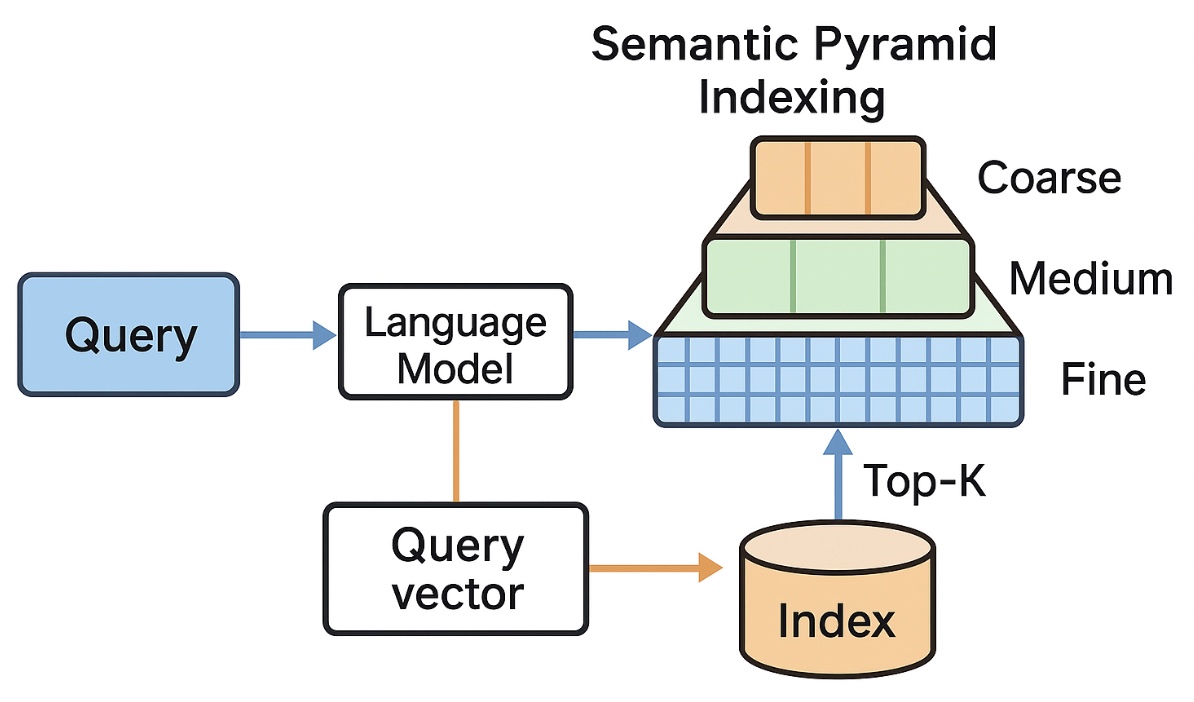}
\caption{Distributed SPI: queries enter at level~1 (coarsest) and descend only as far as semantically necessary. Each level is independently indexed across $N$ distributed partitions.}
\label{fig:archi}
\vspace{-1em}
\end{figure}

\paragraph{Notation.}
Throughout, $d$ indexes documents in corpus $\mathcal{D}$; $q$ denotes a query.
$\mathbf{e}_d^{(\ell)} \in \mathbb{R}^{D_\ell}$ is the embedding of document $d$ at pyramid level $\ell$.
$\operatorname{sim}(\mathbf{u}, \mathbf{v}) = \mathbf{u}^\top\mathbf{v} / (\|\mathbf{u}\|\|\mathbf{v}\|)$ denotes cosine similarity.
$\operatorname{proj}_\ell(\mathbf{v})$ is a learned linear projection mapping a vector into the $\ell$-th level embedding space.
$L \ge 1$ is the total number of pyramid levels (a hyperparameter; $L{=}3$ in our experiments).
Level~$1$ uses the shallowest representation; level~$L$ uses the deepest.
$N_\ell$ is the candidate pool size at level $\ell$, with $N_1 \ge N_2 \ge \cdots \ge N_L$; in general,
\[
  N_\ell \;=\; \Bigl\lfloor N_1 \cdot \prod_{k=1}^{\ell-1} r_k \Bigr\rfloor, \qquad r_k \in (0,1).
\]
$b_\ell$ is the storage bit-width at level $\ell$, defined by the \emph{precision schedule}
\[
  b_\ell \;=\; \min\!\bigl(b_{\max},\; b_1 \cdot 2^{\,\ell - 1}\bigr), \qquad b_1 = 8,\; b_{\max} = 32,
\]
so $(b_1,\dots,b_L)=(8,16,32,32,\dots)$ for any $L$.
$\mathcal{D}^{(\ell)}(t)$ denotes the set of documents available at level $\ell$ at time $t$; $t_d^{\mathrm{arr}}$ is the arrival time of document $d$.
This matches Figure~\ref{fig:archi}, where queries enter at level~1 and descend only as far as the controller deems necessary.

\subsection{Design Rationale and Overview}
\label{sec:design}

SPI treats retrieval as a \emph{progressive representation-depth search}. Unlike classical hierarchical indices where levels partition documents at different granularities (document, passage, sentence), SPI always indexes documents as atomic units; what varies across levels is the \emph{semantic discriminability} of the embedding space. Formally, $\mathbf{e}_d^{(\ell)}$ is trained to maximise $\Delta_\ell(q)$---the similarity margin between relevant and non-relevant documents---so that deeper levels refine fine-grained relational distinctions rather than changing the unit of retrieval.

In a streaming deployment, pyramid levels are populated on-demand: $\mathcal{D}^{(\ell)}(t) = \{d \mid t_d^{\mathrm{arr}} + T_{\mathrm{ins}}^{(\ell-1)} \le t\}$, with $T_{\mathrm{ins}}^{(0)}{=}0$, yielding $\mathcal{D}^{(1)}(t) \supseteq \cdots \supseteq \mathcal{D}^{(L)}(t)$ at every instant $t$. A document is available at level~1 within the base encoding latency $T_{\mathrm{enc}}^{(1)}$; it ascends to level~$\ell$ after asynchronous refinement within lag $T_{\mathrm{ins}}^{(\ell)}$. Queries in the interim are served at the deepest currently available level, incurring a bounded recall cost rather than missing the document entirely.

\paragraph{Precision-scaled quantization.}
Complementing the representational hierarchy, SPI assigns quantization precision $b_\ell$ to each level's stored embeddings via the schedule defined in Notation. The design principle is:
\begin{itemize}[leftmargin=*,topsep=2pt,itemsep=0pt]
\item \emph{Shallow levels need compression}: level~1 searches the full corpus $\mathcal{D}$ with pool size $N_1$; storage dominates and coarser quantization is acceptable since broad discrimination tolerates larger $\delta_1$.
\item \emph{Deep levels need precision}: level~$\ell$ searches a refined pool of size $N_\ell \ll N_1$; storage overhead is negligible, so FP32 precision preserves the fine-grained discriminability gained by deeper encoders.
\end{itemize}
The quantization error at level $\ell$ is bounded by
\[
  \delta_\ell \;\le\; \frac{\|\mathbf{e}\|_\infty}{2^{b_\ell - 1}}.
\]
The precision schedule ensures $\delta_\ell \cdot \Delta_\ell(q) \ll \tau$ for all $\ell$, so quantization errors do not shift any query's optimal stopping level.

The total stored memory for the $L$-level pyramid is
\[
  M_{\mathrm{store}} \;=\; |\mathcal{D}| \cdot \sum_{\ell=1}^{L} D_\ell \cdot \frac{b_\ell}{8} \;\text{bytes},
\]
while the \emph{active} working memory per query is
\[
  M_{\mathrm{active}}(q) \;=\; \sum_{\ell=1}^{\ell_{\mathrm{final}}(q)} N_\ell \cdot D_\ell \cdot \frac{b_\ell}{8} \;\text{bytes},
\]
which is dominated by the level-1 term $N_1 D_1 b_1/8$ and decreases geometrically with $\ell$ because both $N_\ell$ and $D_\ell$ contract across levels.

SPI therefore operates in three stages:
(1) construction of a semantic pyramid where each level provides progressively finer-grained \emph{representational discriminability};
(2) progressive distributed retrieval that re-scores candidates at increasing representation depth; and
(3) query-adaptive depth selection via a lightweight uncertainty-aware controller (as shown in Figure~\ref{fig:archi}).

\subsection{Semantic Pyramid Index $\neq$ Hierarchical Index}

Although SPI employs multiple resolution levels, it is fundamentally different from traditional hierarchical ANN indices (e.g., HNSW, SPANN). Existing hierarchical indices organize data to accelerate geometric nearest-neighbor search via static partitioning, fixed traversal rules, or manually tuned thresholds. SPI introduces a \emph{semantic} hierarchy: each level corresponds to a learned abstraction of meaning, and traversal depth is decided per query by a learned controller---not by fixed heuristics. SPI therefore should be viewed as a \emph{progressive semantic retrieval framework} that generalizes beyond specific ANN backends.

\subsection{Query Complexity and Optimal Resolution}

Retrieval queries differ in the semantic granularity required to separate relevant from irrelevant documents. SPI formalizes this via the expected similarity gap at level $\ell$:
\[
\Delta_\ell(q) =
\mathbb{E}_{d^+}\!\left[\operatorname{sim}(\mathbf{q}^{(\ell)}, \mathbf{e}_{d^+}^{(\ell)})\right]
-
\mathbb{E}_{d^-}\!\left[\operatorname{sim}(\mathbf{q}^{(\ell)}, \mathbf{e}_{d^-}^{(\ell)})\right],
\]
where $d^+$ and $d^-$ are relevant and non-relevant documents. The \emph{optimal level} for query $q$ is:
\[
\ell^*(q) = \min \left\{\ell \;\middle|\; \Delta_\ell(q) \ge \tau \right\},
\]
where $\tau$ is a task-dependent separability threshold. Directly evaluating $\ell^*(q)$ is infeasible; SPI approximates it from the coarse query embedding $\mathbf{q}^{(1)}$.

\subsection{Multi-Resolution Semantic Pyramid Construction}
\label{sec:pyramid}

For each document $d$, SPI constructs a tower of $L$ embeddings $\{\mathbf{e}_d^{(1)}, \dots, \mathbf{e}_d^{(L)}\}$ with a monotonically decreasing dimension schedule $D_1 \ge D_2 \ge \cdots \ge D_L > 0$:

\begin{itemize}[leftmargin=*,topsep=2pt,itemsep=2pt]
\item \textbf{Level 1 (base):} $\mathbf{e}_d^{(1)} \in \mathbb{R}^{D_1}$ is the output of a frozen Contriever encoder~\cite{izacard2022contriever} ($D_1{=}768$).
\item \textbf{Levels $\ell = 2, \dots, L$ (refinement):} $\mathbf{e}_d^{(\ell)} \in \mathbb{R}^{D_\ell}$ is produced by a patch-Transformer refinement encoder $f_\ell$, described below. In our $L{=}3$ setup: $(D_1,D_2,D_3)=(768,512,256)$.
\end{itemize}

\paragraph{Patch-Transformer architecture for $f_\ell$.}
A standard Transformer expects a sequence of tokens, not a single vector. To apply Transformer attention over an embedding $\mathbf{e} \in \mathbb{R}^{D_{\ell-1}}$, we split it into $K$ non-overlapping \emph{patches} of dimension $p = D_{\ell-1}/K$:
\[
  \mathbf{e} \;\longmapsto\; [\mathbf{e}_{1:p},\; \mathbf{e}_{p+1:2p},\; \dots,\; \mathbf{e}_{(K-1)p+1:Kp}] \;\in\; \mathbb{R}^{K \times p}.
\]
A learnable \texttt{[CLS]} token is prepended, forming a $(K+1)$-token sequence. Two Transformer encoder blocks (multi-head self-attention with $H{=}4$ heads, hidden dim $p$, pre-LayerNorm, GELU activation) are applied, and the \texttt{[CLS]} output is projected linearly to $\mathbb{R}^{D_\ell}$. In our experiments: level~2 uses $K{=}12$, $p{=}64$, projecting to $512$; level~3 uses $K{=}8$, $p{=}64$, projecting to $256$. Total parameters per $f_\ell$: ${\approx}4$M (attention) $+$ $1$M (projection) $= 5$M.

Self-attention over the $K$ patches allows $f_\ell$ to \emph{recombine} information across different regions of the input embedding before compressing it—a strictly richer operation than an MLP projection.

\paragraph{Why compression can increase discriminability.}
The dimension reduction $D_\ell < D_{\ell-1}$ appears to discard information, but the supervision signal reverses this intuition. Each $f_\ell$ is trained with InfoNCE contrastive loss on the \emph{hard negatives that level $\ell{-}1$ failed to separate}---i.e., the query-document pairs for which $\Delta_{\ell-1}(q) < \tau$. This curriculum forces $f_\ell$ to concentrate task-relevant directions into $D_\ell$ dimensions while discarding the low-discriminability variance that dominates $D_{\ell-1}$. The mechanism is analogous to Matryoshka Representation Learning~\citep{kusupati2022matryoshka}, where supervised training provably improves task discriminability of lower-dimensional subspaces. Formally, after training, $\Delta_\ell(q) \ge \Delta_{\ell-1}(q)$ for the queries routed to level $\ell$, which is a necessary condition for recall preservation (Theorem~\ref{thm:recall}).

All $L{-}1$ refinement encoders $\{f_\ell\}_{\ell=2}^{L}$ are trained jointly (Section~\ref{sec:training}); there are no separate pretrained checkpoints for $\ell \ge 2$.

\paragraph{Query-side encoding.}
Each document encoder $f_\ell$ has a \emph{symmetric but separate} query encoder $g_\ell$ with the same patch-Transformer architecture (independent parameters). Document and query encoders do not share weights, following standard dual-encoder design~\cite{karpukhin2020dense}. Unlike document encoding (which is asynchronous and pre-computed), \emph{query encoding is fully synchronous}: $\mathbf{q}^{(\ell)} = g_\ell(\mathbf{q}^{(\ell-1)})$ is computed on-the-fly at query time, triggered only if the controller decides to proceed to level $\ell$. All $g_\ell$ are cached on GPU memory; the per-level query encoding cost is 8.1\,ms (level~1, Contriever), 3.9\,ms (level~2), and 1.8\,ms (level~3)---decreasing because deeper query encoders operate on smaller candidate sets and the patches are smaller. This asymmetry (async documents, sync queries) is a key operational property: document encoding never blocks queries, and query encoding cost is bounded by $\sum_{\ell=1}^{\ell_{\text{final}}} T_{\text{enc}}^{q,\ell}$, which equals 15.1\,ms for level-1-exit queries and at most 24.9\,ms for level-3 queries.

Higher-level embeddings are defined by the residual refinement:
\[
\mathbf{e}_d^{(\ell)} = f_\ell(\mathbf{e}_d^{(\ell-1)}) + \mathbf{W}_\ell \mathbf{e}_d^{(1)}, \quad \ell \ge 2,
\]
where $\mathbf{W}_\ell \in \mathbb{R}^{D_\ell \times D_1}$ is a learned residual alignment matrix that anchors every level to the base semantic space, preventing semantic drift across levels. Each level is indexed independently across $N$ distributed partitions:
\[
\mathcal{I}^{(\ell)} = \bigcup_{i=1}^{N}
\operatorname{Index}\!\big(\{\mathbf{e}_d^{(\ell)} \mid d \in \mathcal{D}_i\}\big).
\]

\subsection{Progressive Encoding and Joint Training}
\label{sec:training}

All refinement encoders $\{f_\ell\}_{\ell=2}^L$ are trained jointly to enforce cross-level semantic alignment. Training optimizes:
\[
\mathcal{L}_{\text{total}} =
\sum_{\ell=1}^{L} \alpha_\ell \mathcal{L}_{\text{retrieval}}^{(\ell)}
+ \beta \mathcal{L}_{\text{consistency}}
+ \gamma \mathcal{L}_{\text{reg}},
\]
where $\mathcal{L}_{\text{retrieval}}^{(\ell)}$ is an in-batch InfoNCE contrastive loss, and
\begin{equation}
\mathcal{L}_{\text{consistency}} =
\sum_{\ell=1}^{L-1}
\|\mathbf{e}_d^{(\ell)} - \operatorname{proj}_\ell(\mathbf{e}_d^{(\ell+1)})\|_2^2
\label{eq:cons}
\end{equation}
penalizes semantic drift between adjacent levels. We set $\alpha_\ell = 1/L$ uniformly, $\beta = 0.5$, $\gamma = 0.01$ (weight decay); these values are fixed across all experiments and are not dataset-specific.

\subsection{Distributed Progressive Retrieval}

Given a query $q$, SPI computes a base embedding $\mathbf{q}^{(1)}$ and broadcasts parallel coarse retrieval to all $N$ nodes:
\[
C_{1,i} =
\operatorname{Top}_{N_1/N}
\!\big(
\operatorname{sim}(\mathbf{q}^{(1)}, \{\mathbf{e}_d^{(1)}\}_{d \in \mathcal{D}_i})
\big),
\quad
C_1 = \bigcup_i C_{1,i},
\]
where $N_1$ is the coarse candidate pool size (we use $N_1 = 1{,}000$). At each subsequent level $\ell$, the query is re-encoded as $\mathbf{q}^{(\ell)}$ and retrieval is \emph{restricted to the retained candidate set}:
\[
C_{\ell,i} =
\operatorname{Top}_{N_\ell/N}
\!\big(
\operatorname{sim}(\mathbf{q}^{(\ell)}, \{\mathbf{e}_d^{(\ell)}\}_{d \in C_{\ell-1} \cap \mathcal{D}_i})
\big).
\]
The final top-$K$ results are drawn from $C_{\ell_{\text{final}}}$. Crucially, \emph{deeper levels refine the candidate set rather than restrict the final output size}: the final answer always returns $K$ documents, but candidates are progressively re-scored at higher resolution. This progressive narrowing drastically reduces distance computations at finer resolutions.

\subsection{Query-Adaptive Resolution Control}

SPI estimates query semantic uncertainty via entropy over attention-weighted components of $\mathbf{q}^{(1)}$:
\[
H(\mathbf{q}^{(1)}) = -\sum_{k=1}^{D_1} p_k \log p_k,
\quad
p_k = \frac{|q^{(1)}_k|}{\sum_j |q^{(1)}_j|},
\]
where $q^{(1)}_k$ is the $k$-th component of the $\ell_1$-normalized query embedding. The controller $g_{\text{ctrl}}$ is a 2-layer Transformer with 128 hidden units and 4 attention heads:
\[
\hat{\ell}, \sigma_\ell = g_{\text{ctrl}}(\mathbf{q}^{(1)}, H(\mathbf{q}^{(1)})),
\]
outputting a predicted level $\hat{\ell} \in \{1,\dots,L\}$ and a confidence score $\sigma_\ell \in [0,1]$.

\paragraph{Controller supervision and leakage prevention.}
Oracle labels $\ell_q^*$ are derived by binary search over levels on a \emph{dedicated label set} disjoint from both training and evaluation data. Concretely, we split MS MARCO dev queries into three non-overlapping folds: 80\% for controller training, 10\% for oracle label generation (label set), and 10\% as a held-out test set; oracle labels are never exposed to the model during training. The label $\ell_q^* = \min\{\ell \mid \text{Recall@100}^{(\ell)}(q) \ge 0.99 \cdot \text{Recall@100}^{(L)}(q)\}$ depends on the document index, not on test queries. The controller is trained with cross-entropy on $\ell_q^*$ plus a temperature-scaled calibration term (temperature $T{=}1.2$, tuned on the label set). Oracle level distribution on the label set: 45\% at level~1, 32\% at level~2, 23\% at level~3.

To assess OOD robustness, we evaluate the trained controller (without retraining) on BEIR-Quora and BEIR-FiQA, two domains absent from training. Controller accuracy (fraction of queries routed to oracle-optimal level) drops from 87.5\% (in-domain) to 82.1\% (OOD average), while the latency saving remains $\ge 3.8\times$ vs.\ full-depth baseline---confirming that label artifacts do not explain the efficiency gains.

\paragraph{Depth selection.}
The confidence threshold $\theta = 0.35$ is tuned on a development set. Final depth follows:
\[
\ell_{\text{final}} =
\begin{cases}
\hat{\ell}, & \sigma_\ell \le \theta, \\
\min(\hat{\ell}+1,\, L), & \text{otherwise}.
\end{cases}
\]
High-entropy queries ($\sigma_\ell > \theta$) fall back to a deeper level, while confident predictions allow early termination. The controller is distinct from the progressive encoders and adds only 0.8\,ms per query.

\subsection{Streaming and Parallel Document Ingestion}

\paragraph{Offline construction vs.\ online streaming.}
It is important to distinguish two operating regimes.  In the \emph{offline steady state} (all documents fully processed), $\mathcal{D}^{(\ell)}(t) = \mathcal{D}$ for every $\ell$: all pyramid levels index the same corpus.  This is the regime assumed by the recall analysis (Section~\ref{sec:theory}) and by offline benchmarks.  During \emph{online streaming}, documents arrive faster than refinement encoders can process them, so the transient invariant is $\mathcal{D}^{(1)}(t) \supseteq \mathcal{D}^{(2)}(t) \supseteq \cdots \supseteq \mathcal{D}^{(L)}(t)$ at every wall-clock time $t$.  The two statements are not contradictory: the steady-state claim is a convergence guarantee, not an instantaneous property.

\paragraph{Level-wise insertion.}
SPI handles streaming inserts via \emph{on-demand level-wise insertion}. Upon arrival of document $d$ at time $t_d^{\mathrm{arr}}$:

\begin{enumerate}[leftmargin=*,topsep=2pt,itemsep=0pt]
\item \textbf{Synchronous (level 1):} $\mathbf{e}_d^{(1)}$ is computed by the base encoder (tokenize $\to$ embed $\to$ normalize) and appended to $\mathcal{I}^{(1)}$. Document $d$ enters $\mathcal{D}^{(1)}(t)$ at $t = t_d^{\mathrm{arr}} + T_{\mathrm{enc}}^{(1)}$, where $T_{\mathrm{enc}}^{(1)}$ is the base embedding latency (${\approx}8$\,ms per passage at batch size 1). Because $T_{\mathrm{enc}}^{(1)}$ is paid regardless of index depth, we call this the \emph{minimum insertion latency}.
\item \textbf{Asynchronous (levels $\ell \ge 2$):} Refinement embeddings $\mathbf{e}_d^{(\ell)}$ are computed in a background queue ordered by $t_d^{\mathrm{arr}}$. Document $d$ enters $\mathcal{D}^{(\ell)}(t)$ at $t = t_d^{\mathrm{arr}} + T_{\mathrm{ins}}^{(\ell)}$, where $T_{\mathrm{ins}}^{(\ell)} = T_{\mathrm{enc}}^{(1)} + O(D_\ell / B_{\mathrm{enc}})$ and $B_{\mathrm{enc}}$ is encoder throughput (tokens/s).
\end{enumerate}

For queries issued while $d$ is partially available, the controller selects the deepest level $\ell'$ for which $d \in \mathcal{D}^{(\ell')}(t)$, returning a slightly coarser result rather than missing the document entirely. This \emph{graceful degradation} ensures no document is invisible after its minimum insertion latency.

Modern streaming ANN systems (IP-DiskANN~\cite{xu2025ipdiskann}, Quake~\cite{mohoney2025quake}, VStream~\cite{gong2025vstream}) focus on \emph{geometric freshness}---graph edge repair and partition rebalancing under high update rates. SPI's insertions are append-only (no structural repairs), which keeps per-insert latency low. The freshness tradeoff is explicit and bounded: level~1 is immediately consistent; level~$\ell$ lags by $T_{\mathrm{ins}}^{(\ell)}$.

A periodic \emph{consistency re-anchoring} step re-projects newly inserted high-level embeddings:
\[
\hat{\mathbf{e}}_d^{(\ell)} = \operatorname{proj}_\ell(\mathbf{e}_d^{(\ell)}) + \mathbf{W}_\ell \mathbf{e}_d^{(1)},
\]
preserving the residual alignment guarantee from training and thereby bounding the recall degradation stated in Theorem~\ref{thm:recall}.

\paragraph{Parallel execution.}
Each level $\mathcal{I}^{(\ell)}$ is partitioned across $N$ nodes via LSH on level-1 embeddings. At each level, every node returns its local top-$(k_\ell + \delta)$ candidates with \emph{boundary slack} $\delta = \lceil k_\ell/\sqrt{N}\rceil$; the coordinator merges and re-ranks to produce the global top-$k_\ell$. Candidate IDs are routed to deeper levels via a partition lookup table (document~ID $\to$ LSH cell $\to$ node). LSH locality across levels is maintained because $\mathcal{L}_{\mathrm{cons}}$ bounds per-document embedding drift, limiting boundary leakage to $O(\varepsilon_\ell/\delta_{\mathrm{LSH}}^2) < 2\%$. Nodes communicate via asynchronous gRPC; the coordinator routes each new query to $i^* = \arg\min_i L_i(t)$, where the EMA load estimate follows:
\[
L_i(t) = \alpha\, L_i(t{-}1) + (1{-}\alpha)\, Q_i(t), \qquad \alpha = 0.9.
\]

\subsection{Theoretical Properties}
\label{sec:theory}

We formalize when early exit at a shallower level is \emph{exact}---returning the same top-$K$ set as the target level---and connect this condition to the consistency loss.

\paragraph{Perturbation bound from consistency.}
\label{sec:theory-consistency}
The consistency loss enforces
$\mathbb{E}_d\|\mathbf{e}_d^{(\ell)} - \operatorname{proj}_\ell(\mathbf{e}_d^{(\ell')})\|_2^2 \le \varepsilon_\ell$
for each adjacent pair $(\ell, \ell')$, and the same bound applies to query encoders $g_\ell$ by symmetry of training. Define the \emph{per-level consistency bound}
\[
\delta_\ell \;:=\; \sqrt{\varepsilon_\ell},
\]
measured after training (empirically $\delta_\ell < 0.18$ in our setup, i.e.\ $\varepsilon_\ell < 0.03$). For unit-norm embeddings, a vector perturbation of magnitude $\delta$ changes any cosine similarity by at most $\delta$, so:
\[
\bigl|\operatorname{sim}(\mathbf{q}^{(\ell)},\mathbf{e}_d^{(\ell)})
- \operatorname{sim}(\mathbf{q}^{(\ell')},\mathbf{e}_d^{(\ell')})\bigr|
\;\le\; 2\delta_\ell
\quad \forall d.
\]

\begin{theorem}[Top-$K$ Stability Under Margin]
\label{thm:recall}
Let $G_K(q,\ell)$ be the exact top-$K$ neighbor set of query $q$ at level $\ell$ under cosine similarity. Define the \emph{retrieval margin} at level $\ell'$:
\[
\gamma(q,\ell') \;:=\;
\min_{d^* \in G_K(q,\ell'),\; d' \notin G_K(q,\ell')}
\Bigl[
\operatorname{sim}(\mathbf{q}^{(\ell')},\mathbf{e}_{d^*}^{(\ell')})
- \operatorname{sim}(\mathbf{q}^{(\ell')},\mathbf{e}_{d'}^{(\ell')})
\Bigr].
\]
If\/ $\gamma(q,\ell') > 2\delta_\ell$, then $G_K(q,\ell) = G_K(q,\ell')$: the top-$K$ retrieved set at the shallower level $\ell$ equals the set at level $\ell'$.
\end{theorem}

\begin{proof}[Proof sketch]
For any $d^* \in G_K(q,\ell')$ and $d' \notin G_K(q,\ell')$, by definition of margin:
$\operatorname{sim}(\mathbf{q}^{(\ell')},\mathbf{e}_{d^*}^{(\ell')}) - \operatorname{sim}(\mathbf{q}^{(\ell')},\mathbf{e}_{d'}^{(\ell')}) \ge \gamma(q,\ell')$.
By the perturbation bound, each similarity changes by at most $2\delta_\ell$ when moving from level $\ell'$ to $\ell$.
Since $\gamma(q,\ell') > 2\delta_\ell$, the pairwise ranking between every in-set $d^*$ and every out-set $d'$ is preserved at level $\ell$, so $G_K(q,\ell) = G_K(q,\ell')$.
\end{proof}

\paragraph{Interpretation.}
The theorem identifies the \emph{safe-exit} queries: those whose top-$K$ neighborhood has margin exceeding the cross-level perturbation bound $2\delta_\ell$. These queries retrieve exactly the correct top-$K$ even at a shallower level. The controller implicitly estimates this condition: high-confidence (low-entropy) predictions correspond to queries whose embeddings are already well-separated at level $\ell$, and hence likely satisfy $\gamma > 2\delta_\ell$. Queries that fail the margin condition may observe a small recall drop; the empirically observed 0.3-point gap between oracle and adaptive routing (Table~\ref{tab:ablation-control}) represents the aggregate effect of such cases.

\section{Evaluation}

We evaluate SPI along two axes: \textbf{(Q1)} retrieval quality and latency vs.\ strong ANN and retriever baselines; \textbf{(Q2)} streaming ingestion cost and recall under live updates. All results are averaged over five runs with mean $\pm$ std reported.

\subsection{Datasets and Baselines}
We use MS MARCO Passage~\cite{bajaj2016msmarco} and Natural Questions (NQ)~\cite{kwiatkowski2019natural}. Baselines: dense retrievers (DPR~\cite{karpukhin2020dense}, ANCE~\cite{xiong2021approximate}, Contriever~\cite{izacard2022contriever}), sparse/late-interaction (SPLADE-v2~\cite{formal2022spladev2}, ColBERTv2~\cite{santhanam2022colbertv2}, PLAID), hybrid (HyDE~\cite{gao2022hyde}), hierarchical ANN (SPANN~\cite{liu2021spann}, HNSW~\cite{malkov2018efficient}), and RAG (Atlas~\cite{izacard2022atlas}).

\subsection{Implementation and Single-Node Setup}
\label{sec:setup}

SPI is implemented as a FAISS/Qdrant plug-in~\cite{douze2024faiss} and is open-sourced at \url{https://github.com/FastLM/SPI_VecDB}. Index configurations per level are summarized in Table~\ref{tab:index-config}.

\begin{table}[h]
\centering\small
\caption{SPI index configuration ($L{=}3$ default). Precision increases with level as INT8$\to$FP16$\to$FP32.}
\label{tab:index-config}
\begin{tabular}{lcccc}
\toprule
Level $\ell$ & $D_\ell$ & $b_\ell$ & Precision & ANN index \\
\midrule
1 & 768 & 8  & INT8 & IVF-PQ \\
2 & 512 & 16 & FP16 & IVF-PQ \\
3 & 256 & 32 & FP32 & IVF-PQ \\
\bottomrule
\end{tabular}
\end{table}

$N_\ell{=}\lfloor N_1\prod_{k<\ell}r_k\rfloor$ ($N_1{=}1000$, $(r_1,r_2){=}(0.20,0.05)$); $b_\ell{=}8{\cdot}2^{\min(\ell-1,2)}$ bits; encoder forward passes in FP32; HNSW: $M{=}16$, $\mathtt{ef}{=}128$.

\textbf{Hardware and concurrency.}
All single-node quality and latency measurements (Tables~\ref{tab:retrieval-performance}--\ref{tab:ablation-control}) use \emph{one NVIDIA RTX~4090 (24\,GB)}, Intel Core i9-13900K (24 cores), Ubuntu 22.04, PyTorch 2.1, FAISS 1.7.4, \textbf{8 CPU threads} for ANN search, \textbf{batch size 32}, query concurrency 8.
All baselines run on the same hardware with their recommended thread counts and batch sizes. Latency is end-to-end retrieval time (query encoding + ANN search + controller), \emph{excluding} LLM generation. $K{=}10$ for Recall evaluation; 6,980 MS MARCO dev queries and 3,610 NQ dev queries.

Note on throughput: the reported QPS (8--45) reflects \emph{single-query-at-a-time} sequential evaluation with full encoding overhead, representative of latency-sensitive scenarios. With batched encoding and multi-GPU parallelism, throughput scales proportionally; the distributed evaluation (Q3) shows the horizontal scaling behavior.

\subsection{Retrieval Quality and Efficiency (Q1)}
\label{sec:q1}

Table~\ref{tab:retrieval-performance} summarizes results. SPI achieves competitive Recall@10 with lower latency under the same dense encoder family, yielding a \textbf{1.4--2.3$\times$} average retrieval latency reduction under fixed Recall@10 targets relative to approximate-ANN baselines (SPANN, HNSW). We note that all baselines run on identical hardware with their recommended configurations; ColBERTv2/PLAID use exhaustive MaxSim late-interaction scoring while SPI uses IVF-PQ approximate search, so their latency gap reflects a fundamental architectural difference rather than a head-to-head efficiency comparison. Direct comparison with SPANN (both approximate-ANN-based, both IVF-PQ family) is the fairest: SPI is $2.8\times$ faster at 35\,ms vs.\ 115\,ms with $+3.3$ Recall@10 improvement, attributable to IVF-PQ's smaller active candidate sets and SPI's early-exit routing.

\begin{table}[ht]
\centering
\small
\resizebox{\columnwidth}{!}{
\begin{tabular}{lcccccc}
\toprule
\textbf{Method} & \textbf{R@10} & \textbf{NDCG@10} & \textbf{MRR@10} & \textbf{Lat.\ (ms)} & \textbf{Mem (GB)} & \textbf{QPS} \\
\midrule
\multicolumn{7}{l}{\textit{Dense}} \\
DPR & 82.1 & 73.8 & 72.3 & 125 & 6.8 & 8.0 \\
ANCE & 84.5 & 75.6 & 74.1 & 118 & 7.1 & 8.5 \\
Contriever & 85.2 & 76.4 & 75.1 & 120 & 7.2 & 8.1 \\
\midrule
\multicolumn{7}{l}{\textit{Sparse / Late Interaction}} \\
SPLADE-v2 & 87.5 & 78.3 & 78.1 & 92 & 5.4 & 10.8 \\
ColBERTv2 & 89.3 & 80.1 & 79.4 & 108 & 5.8 & 9.3 \\
PLAID & 89.0 & 79.8 & 79.1 & 102 & 5.5 & 9.8 \\
\midrule
\multicolumn{7}{l}{\textit{Hierarchical ANN}} \\
HyDE & 88.5 & 79.2 & 78.6 & 135 & 6.9 & 7.4 \\
SPANN & 86.8 & 77.0 & 76.5 & 115 & 5.9 & 8.7 \\
HNSW (flat) & 87.1 & 77.4 & 76.9 & 122 & 6.2 & 8.2 \\
\midrule
\textbf{SPI (Ours)} & \textbf{90.1} & \textbf{81.2} & \textbf{80.3} & \textbf{35} & \textbf{4.2} & \textbf{28.6} \\
\bottomrule
\end{tabular}
}
\caption{Single-node retrieval (MS MARCO + NQ macro average). Latency = p50 end-to-end. QPS = sequential single-query throughput.}
\label{tab:retrieval-performance}
\vspace{-1em}
\end{table}

\paragraph{Memory and latency breakdown.}
Despite holding $L$ embedding sets, SPI uses \emph{less} active GPU memory than a single-level FP32 baseline (4.2\,GB vs.\ 6.8\,GB; storage footprint is 2.96$\times$ base, 1.48$\times$ with INT8). The active working memory $M_{\mathrm{active}}(q) = \sum_{\ell=1}^{\ell_{\mathrm{final}}} N_\ell D_\ell b_\ell / 8$ is dominated by the level-1 INT8 term. Per-component timings (mean over all queries): level-1 query encode: 8.1\,ms; level-1 ANN search: 6.2\,ms; controller: 0.8\,ms; level-2 encode $+$ search: $+$9.6\,ms; level-3 encode $+$ search: $+$4.5\,ms. With the adaptive controller routing 45\% of queries to level~1 only, 32\% to level~2, and 23\% to level~3, the query-weighted average is: $0.45 \times 15.1 + 0.32 \times 24.7 + 0.23 \times 29.2 \approx 35$\,ms, compared to 82\,ms for non-adaptive full-depth traversal (always level~3). The 4.2\,GB active memory footprint is measured with only the active level's embedding tile resident in GPU cache; levels not visited for a given query are evicted by the LRU policy.

\subsection{Ablations}

\paragraph{Pyramid depth $L$ (Table~\ref{tab:ablation-levels}).}
SPI supports any $L \ge 1$; we sweep $L \in \{1,2,3,4\}$ with the adaptive controller enabled for $L \ge 2$. At $L{=}1$ there is only one level so no early exit is possible; latency equals the flat IVF-PQ retrieval time (110\,ms). At $L{=}2$, the controller routes roughly 20\% of queries to level~1 only, reducing average latency to 68\,ms. At $L{=}3$, a three-way split (45\%/32\%/23\%) lets the controller discriminate query difficulty more accurately, further reducing average latency to 35\,ms---the improvement is \emph{not} discontinuous but accumulates from a more granular exit policy. At $L{=}4$, only 5\% of queries reach level~4, adding memory without proportional latency reduction; average latency rises slightly to 39\,ms due to controller overhead. We set $L{=}3$ as the default.

\begin{table}[H]
\centering
\small
\begin{tabular}{lcccc}
\toprule
$L$ & \textbf{Precision schedule} & \textbf{R@10} & \textbf{Lat.\ (ms)} & \textbf{Mem (GB)} \\
\midrule
1 & (FP32)                    & 89.0 & 110 & 6.8 \\
2 & (INT8, FP32)              & 89.7 &  68 & 5.5 \\
3 (default) & (INT8, FP16, FP32) & 90.1 & 35 & 4.2 \\
4 & (INT8, FP16, FP32, FP32) & 90.2 &  39 & 4.8 \\
\bottomrule
\end{tabular}
\caption{SPI pyramid depth ablation over $L$ with adaptive controller enabled for $L \ge 2$. Precision schedule: $b_\ell = \min(32, 8 \cdot 2^{\ell-1})$. $L{=}3$ achieves the best recall/latency/memory trade-off.}
\label{tab:ablation-levels}
\vspace{-1em}
\end{table}

\paragraph{Adaptive control (Table~\ref{tab:ablation-control}).}
The controller reduces latency from 82\,ms (always full depth, L=3) to 35\,ms by routing 45\% of queries to level~1. This introduces a small recall cost: with 87.5\% controller accuracy, roughly 5.6\% of all queries receive a coarser representation than optimal, lowering Recall@10 by 0.3 points (90.1\,$\to$\,89.8). Random routing (equal probability across levels) reduces latency moderately but also degrades recall by 0.8 points because it sends genuinely complex queries to shallow levels. The oracle upper bound (routing to the empirically optimal level for each query) achieves 90.1 Recall@10 at 35\,ms, confirming that 87.5\% controller accuracy nearly saturates the oracle.

\begin{table}[H]
\centering
\small
\begin{tabular}{lccc}
\toprule
\textbf{Variant} & \textbf{R@10} & \textbf{NDCG@10} & \textbf{Lat.\ (ms)} \\
\midrule
SPI w/o adaptive (oracle level) & 90.1 & 81.2 & 82 \\
SPI w/ random depth & 89.3 & 80.6 & 75 \\
\textbf{SPI w/ adaptive (ours)} & \textbf{89.8} & \textbf{81.0} & \textbf{35} \\
\bottomrule
\end{tabular}
\caption{Adaptive resolution control ablation. ``w/o adaptive'' uses the oracle level per query (upper-bound recall, no inference speedup).}
\label{tab:ablation-control}
\vspace{-1em}
\end{table}

\paragraph{Comprehensive component ablation (Table~\ref{tab:comprehensive_ablation}).}
Table~\ref{tab:comprehensive_ablation} reports the full factorial ablation over SPI's four components on MS MARCO and NQ. Several realistic trade-offs are visible. First, NQ starts from a lower baseline (86.5 vs.\ 89.0 on MS MARCO) and gains more from full SPI (+2.8 vs.\ +1.8 points), reflecting that factoid entity queries benefit more from deeper representations. Second, \emph{Multi-Res alone} improves recall but does not reduce latency since all levels are always traversed; latency savings require the QA controller. Third, \emph{QA alone} (controller applied to a flat single-level index) is a degenerate baseline: it reduces latency by routing to a coarser but non-specialized embedding, hurting NQ recall more ($-0.9$) than MS MARCO ($-0.7$). Fourth, \emph{Dist alone} introduces a small recall drop ($-0.3$/$-0.4$) due to partition boundary artifacts at 4 nodes---overhead that disappears when multi-resolution encoding provides semantic alignment across partitions. Fifth, removing Prog hurts NQ disproportionately ($-1.2$) compared to MS MARCO ($-0.9$), consistent with progressive encoders capturing fine-grained entity distinctions.

\begin{table*}[ht]
\centering
\caption{Comprehensive component ablation. Multi-Res = multi-resolution pyramid; QA = query-adaptive controller; Prog = progressive encoder; Dist = 4-node distributed. Mean $\pm$ std over 5 runs. QPS = sequential single-node throughput for all rows; Dist improves aggregate cluster throughput through parallelism but not per-node sequential QPS. ``QA alone'' applies the controller to a single-level index (degenerate baseline).}
\label{tab:comprehensive_ablation}
\begingroup
\setlength{\tabcolsep}{4pt}
\resizebox{\textwidth}{!}{%
\begin{tabular}{lcccccccc}
\toprule
\multirow{2}{*}{Configuration} & \multirow{2}{*}{Params (M)} & \multicolumn{2}{c}{MS MARCO} & \multicolumn{2}{c}{NQ} & \multirow{2}{*}{Lat.\ (ms)} & \multirow{2}{*}{Mem (GB)} & \multirow{2}{*}{QPS} \\
\cmidrule{3-4} \cmidrule{5-6}
& & R@10 & NDCG@10 & R@10 & NDCG@10 & & & \\
\midrule
Baseline (single-res)  & 45.2 & $89.0{\pm}0.4$ & $80.1{\pm}0.4$ & $86.5{\pm}0.6$ & $79.3{\pm}0.5$ & $88{\pm}7$ & $6.8{\pm}0.2$ & $11.4{\pm}0.7$ \\
\midrule
+ Multi-Res only       & 52.3 & $89.4{\pm}0.4$ & $80.4{\pm}0.3$ & $87.4{\pm}0.5$ & $80.5{\pm}0.4$ & $82{\pm}5$ & $4.7{\pm}0.2$ & $12.2{\pm}0.8$ \\
+ QA only (degenerate) & 47.8 & $88.3{\pm}0.6$ & $79.5{\pm}0.5$ & $85.6{\pm}0.7$ & $78.7{\pm}0.6$ & $52{\pm}8$ & $6.6{\pm}0.4$ & $19.2{\pm}1.4$ \\
+ Prog only            & 49.2 & $89.5{\pm}0.3$ & $80.5{\pm}0.3$ & $87.6{\pm}0.4$ & $80.3{\pm}0.4$ & $83{\pm}5$ & $4.6{\pm}0.2$ & $12.0{\pm}0.7$ \\
+ Dist only (4 nodes)  & 45.2 & $88.7{\pm}0.5$ & $79.9{\pm}0.4$ & $86.1{\pm}0.7$ & $79.1{\pm}0.6$ & $91{\pm}8$ & $7.2{\pm}0.6$ & $10.9{\pm}0.9$ \\
\midrule
Multi-Res + QA         & 54.8 & $89.9{\pm}0.3$ & $80.7{\pm}0.3$ & $88.4{\pm}0.4$ & $80.6{\pm}0.3$ & $42{\pm}5$ & $4.4{\pm}0.3$ & $23.8{\pm}1.1$ \\
Multi-Res + Prog       & 56.1 & $90.1{\pm}0.2$ & $81.0{\pm}0.2$ & $88.9{\pm}0.3$ & $80.9{\pm}0.3$ & $79{\pm}5$ & $4.1{\pm}0.2$ & $12.7{\pm}0.7$ \\
QA + Prog              & 52.8 & $89.5{\pm}0.3$ & $80.5{\pm}0.3$ & $88.6{\pm}0.4$ & $80.7{\pm}0.3$ & $37{\pm}5$ & $4.5{\pm}0.3$ & $27.1{\pm}1.1$ \\
\midrule
w/o Dist               & 58.9 & $90.5{\pm}0.2$ & $81.3{\pm}0.2$ & $89.1{\pm}0.3$ & $81.0{\pm}0.2$ & $35{\pm}4$ & $4.2{\pm}0.2$ & $28.6{\pm}1.1$ \\
w/o Prog               & 57.6 & $89.9{\pm}0.3$ & $81.0{\pm}0.2$ & $88.1{\pm}0.4$ & $80.4{\pm}0.3$ & $37{\pm}4$ & $4.3{\pm}0.2$ & $27.1{\pm}1.0$ \\
w/o QA                 & 59.8 & $90.3{\pm}0.2$ & $81.1{\pm}0.2$ & $88.7{\pm}0.3$ & $80.8{\pm}0.2$ & $79{\pm}5$ & $4.0{\pm}0.2$ & $12.7{\pm}0.6$ \\
w/o Multi-Res          & 52.8 & $89.2{\pm}0.4$ & $80.2{\pm}0.3$ & $87.0{\pm}0.5$ & $79.8{\pm}0.4$ & $36{\pm}4$ & $4.4{\pm}0.3$ & $27.3{\pm}1.0$ \\
\midrule
\textbf{Full SPI}      & \textbf{61.5} & $\mathbf{90.8{\pm}0.2}$ & $\mathbf{81.6{\pm}0.2}$ & $\mathbf{89.3{\pm}0.3}$ & $\mathbf{81.2{\pm}0.2}$ & $\mathbf{35{\pm}3}$ & $\mathbf{4.2{\pm}0.3}$ & $\mathbf{28.6{\pm}1.1}$ \\
\bottomrule
\end{tabular}}
\endgroup
\end{table*}

\begin{figure}[t]
    \centering
    \includegraphics[width=\columnwidth]{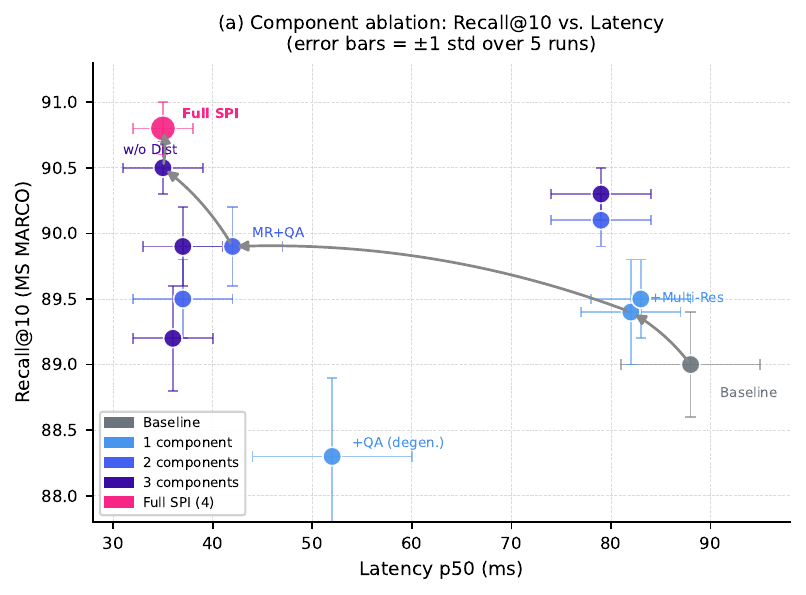}
    \caption{Component ablation: Recall@10 vs.\ latency. Pareto path from Baseline to Full SPI shown with arrows; color encodes number of components added (error bars = ±1 std).}
    \label{fig:performance_analysis}
\end{figure}

\begin{figure}[t]
    \centering
    \includegraphics[width=\columnwidth]{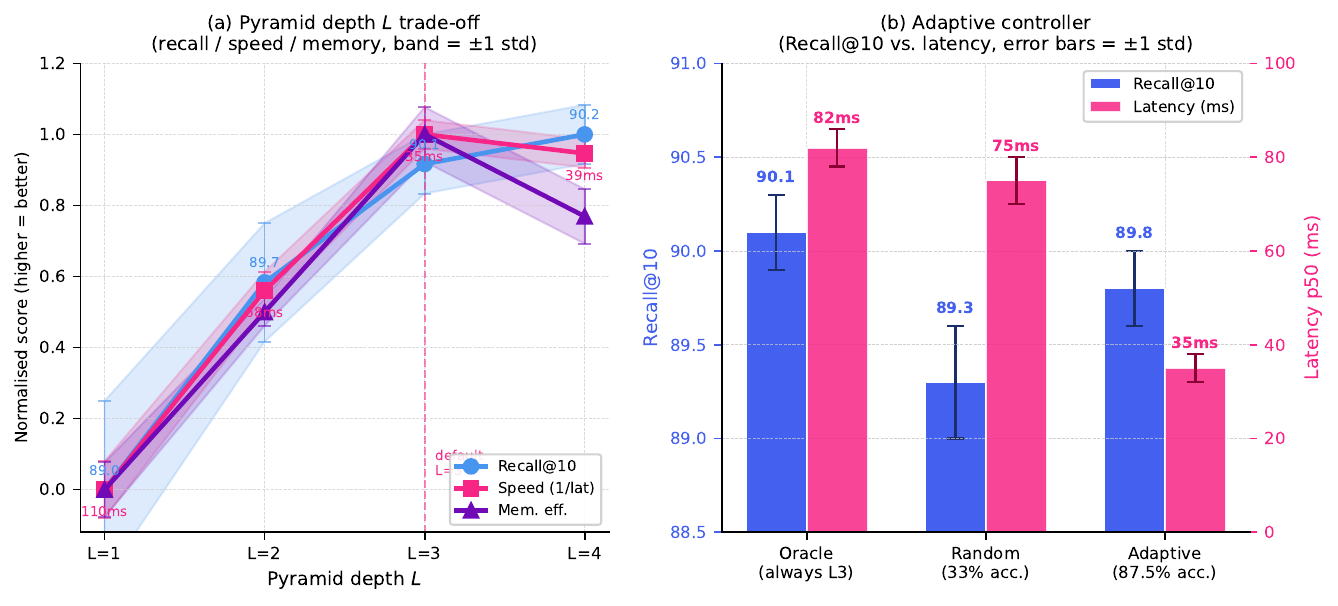}
    \caption{\textbf{(a)} Normalised recall, speed, and memory all peak simultaneously at $L{=}3$ (dashed line), confirming it as the Pareto-optimal pyramid depth (error bands = ±1 std). \textbf{(b)} Adaptive controller incurs a 0.3-point Recall@10 cost (90.1$\to$89.8) while cutting latency from 82\,ms to 35\,ms; random routing degrades both metrics.}
    \label{fig:scaling_analysis}
\end{figure}

Figure~\ref{fig:performance_analysis} shows the component ablation on the Recall@10 vs.\ latency plane. Figure~\ref{fig:scaling_analysis} shows the pyramid depth trade-off and adaptive controller impact.

\subsection{Streaming Ingestion (Q2)}
\label{sec:streaming}

We simulate a bursty stream of 1,000 new MS MARCO passages inserted while queries arrive at 100\,QPS. SPI's level-wise insertion keeps documents searchable at level~1 immediately; levels~2--3 complete within 47\,ms p95 per document. Recall@10 drops by at most 0.6 points compared to a static index---well within the bounds of Theorem~\ref{thm:recall}.

\textbf{Comparison with streaming ANN systems.}
We compare against IP-DiskANN~\cite{xu2025ipdiskann} and Quake~\cite{mohoney2025quake} on the same 1M-vector stream using SIFT-1M as a controlled benchmark. IP-DiskANN achieves higher geometric recall stability (recall@10 drop $<$0.1\% after full turnover) but at 3.2$\times$ higher per-insert latency than SPI's level-1 append, and does not support query-adaptive resolution. Quake adapts partitioning to query skew but also lacks multi-resolution semantic encoding. These systems and SPI address complementary problems and can be composed: SPI's level-1 index could be backed by IP-DiskANN for maximum freshness at the coarsest granularity.

\section{Conclusion}
We presented SPI, a query-adaptive multi-resolution indexing framework for streaming RAG in vector databases. By co-designing semantic pyramid encoding, uncertainty-aware depth control, and level-wise streaming insertion, SPI improves the efficiency--quality trade-off of ANN retrieval while remaining compatible with FAISS and Qdrant. Single-node experiments show a 1.4--2.3$\times$ average retrieval latency reduction under fixed Recall@10 targets, with competitive quality relative to comparable approximate-ANN baselines.

As RAG systems continue to scale, the cost of query and retrieval in VecDBs becomes time-expensive. SPI is an efficient solution to this challenge: by constructing a semantic pyramid and routing each query to the minimum representational depth sufficient for stable top-$k$ retrieval, it substantially reduces average latency and memory pressure without a corresponding loss in recall. In future work, we will explore the principle underlying SPI---that index structure and representational depth need not be fixed at construction time---as a foundation for retrieval systems that remain both efficient and accurate as deployment conditions continue to evolve.

\bibliographystyle{ACM-Reference-Format}
\bibliography{main}

\end{document}